%% file: oagm2013_bader-final.tex
\documentclass[final]{OAGM}
\OAGMarXiv{1304.1876}

\usepackage{graphicx}
\usepackage[utf8x]{inputenc}  
\usepackage{subfigure}
\usepackage{wrapfig}
\usepackage{multirow}
\usepackage[fleqn]{amsmath}

\usepackage{color,xcolor,ucs}
\graphicspath{{}}

\newcommand{\executeiffilenewer}[3]{%
 \ifnum\pdfstrcmp{\pdffilemoddate{#1}}{\pdffilemoddate{#2}}>0{\immediate\write18{#3}}\fi%
}

\title{Visual Room-Awareness for Humanoid Robot Self-Localization}

\author{Markus Bader \and Johann Prankl \and Markus Vincze \\
  Automation and Control Institute, Vienna University of Technology, Austria}

\begin{document}
\maketitle

\begin{abstract}
Humanoid robots without internal sensors such as a compass tend to lose their orientation after a fall.
Furthermore, re-initialisation is often ambiguous due to symmetric man-made environments.
The  room-awareness module proposed here is inspired by the results of psychological experiments and improves existing self-localization strategies by mapping and matching the visual background with colour histograms.
The matching algorithm uses a particle-filter to generate hypotheses of the viewing directions independent of the self-localization algorithm and generates confidence values for various possible poses.
The robot's behaviour controller uses those confidence values to control self-localization algorithm to converge to the most likely pose and prevents the algorithm from getting stuck in local minima.
Experiments with a symmetric Standard Platform League RoboCup  playing field with a simulated and a real humanoid NAO robot show the significant improvement of the system.\end{abstract}

\section{Introduction}
\begin{wrapfigure}{r}{0.3\textwidth}
    \includegraphics[width=0.3\textwidth]{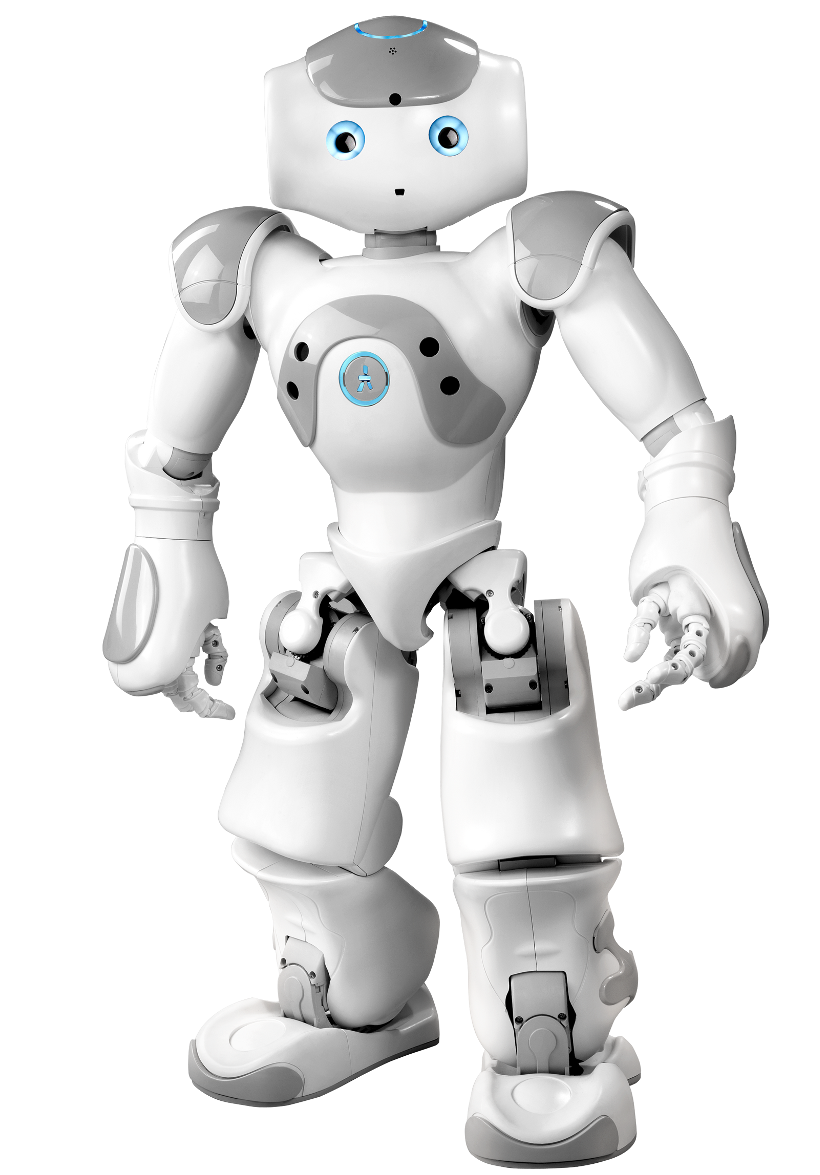}
    \label{fig:Nao}
  \caption{Aldebaran NAO v4, the robot has no electronic compass.}
\end{wrapfigure}

\begin{figure}
\includegraphics[width=\textwidth]{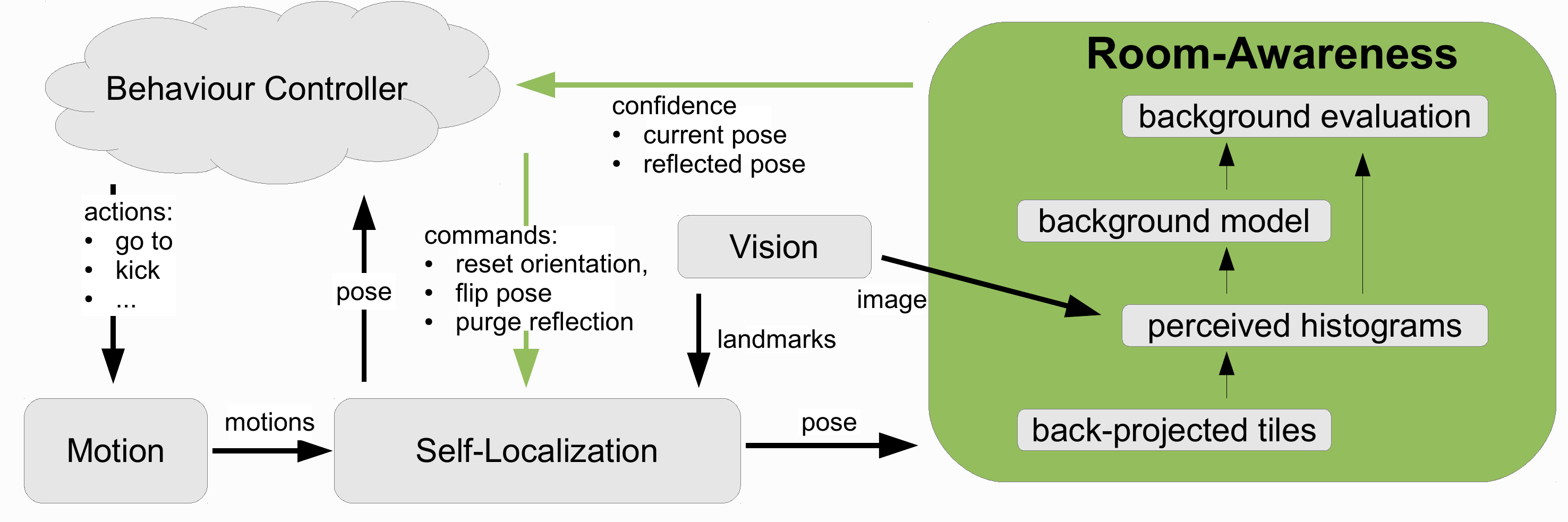}
\caption{The room-awareness module, its integration, in- and output and internal work-flow.}
\label{fig:TheRoomAwarenessModule}
\end{figure}

Rule changes in 2012 in the RoboCup \textit{Standard Platform League} (SPL) resulted in a symmetrical playing field with identical goals. 
Therefore, participating teams must keep track of the direction of play during the game to prevent own goals. 
This poses a problem because sensors like an electronic compass are not allowed. Figure~\ref{fig:Nao} shows the humanoid robot used for the experiments.
Here, the  approach presented in order to confront this challenge is inspired by psychological beliefs and recent experimental results with symmetric environments~\cite{Hermer1994},~\cite{Huttenlocher2007} and~\cite{Lee2012} which proved the existence of spontaneous reorientation in animals and humans.
These experiments showed that the geometric structure of a room has a strong influence on reorientation capabilities.
Lee~et~al.~\cite{Lee2012} proved that the geometric impression of a room can be altered by using  2D shapes (dots of two sizes) printed on walls, supporting or suppressing the subjective geometric impression. 
Three-year-old children spontaneously reoriented in these experiments to the correct corner when the larger-sized dots were on the longer wall of a rectangular room, emphasising the 3D impression.
The influence of the geometric impression of a space on the ability to reorient, as proven in the aforementioned psychological experiments, was integrated into the approach here through the mapping of the visual background of the robot's surroundings beyond the known playing field.
This is done by using colour histograms in order to keep track of the robot's orientation in the room-awareness module, which estimates the robot's orientation, as a separate module. This  enables the computation of independent confidence values for the current pose generated by the self-localization module. 
Fig.~\ref{fig:TheRoomAwarenessModule} shows the integration and the separation of the room-awareness and self-localization module. 
The robot's Behaviour Controller (BC) uses these confidence values to supervise the self-localization module with three simple commands:
\begin{itemize}
\item \textbf{flip pose} \\
This command changes the robot's belief of being in a symmetric reflection.
\item \textbf{purge reflection} \\
This command triggers the self-localization algorithm to remove pose beliefs in symmetric reflections. 
\item \textbf{reset orientation}\\
This command resets the self-localization algorithm's belief in the robot's orientation. The robot's position belief remains untouched. It is used after the internal sensors detect a fall.
\end{itemize}
Therefore, the BC is able to optimise the particle filter used by the self-localization algorithm \cite{Laue2007}, \cite{Thrun2005}. 
A wrong particle cluster, for example, is removed using the \textit{purge reflection} signal if the confidence of the current pose wins over the reflected pose confidence (as in Fig.~\ref{fig:RoomAwarenessOnImage}), or a \textit{flip pose} will be triggered if the reflected pose wins. 
The scientific contributions of this paper are a room-awareness module which mimics a human-like belief of one’s orientation and its integration, allowing the robot's BC to trigger a spontaneous reorientation. 
\\The next section presents related work on the topic followed by a detailed description of the approach and supporting results in Section~\ref{sec:Results}. The results are summarised in Section~\ref{sec:Conclusion}. 

\section{Related Work}\label{sec:RelatedWork}
The strategies presented by different RoboCup-Teams during the past year can be classified into two main groups.
\begin{itemize}
\item Strategies using \textbf{non-static features on the playing field} \newline
Recognized objects like the game ball or other robots, which are needed to play soccer, are used as non-static features in order to overcome the symmetry problem.
\item Strategies using \textbf{features beyond the playing field} \newline
New features in the background are used in addition to features on the symmetric playing field.
\end{itemize}
\begin{figure}
        \centering
        \subfigure[This figure shows two possible valid robot poses.]
        {
            \includegraphics[width=0.3\textwidth]{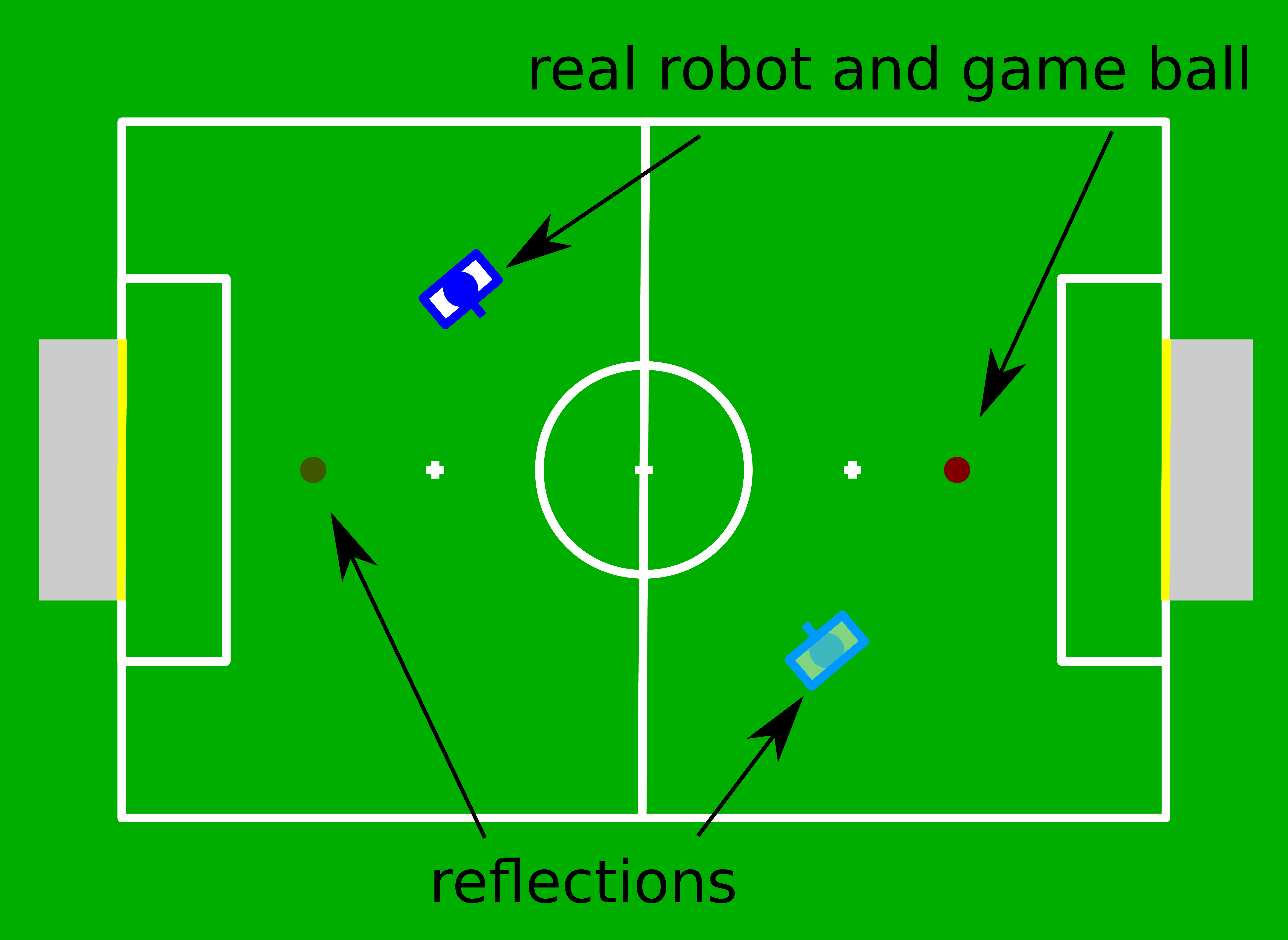}
            \label{fig:TwoPossiblePoses}
        }
        \subfigure[Real robot view of opponent goal.]
        {
            \includegraphics[width=0.3\textwidth]{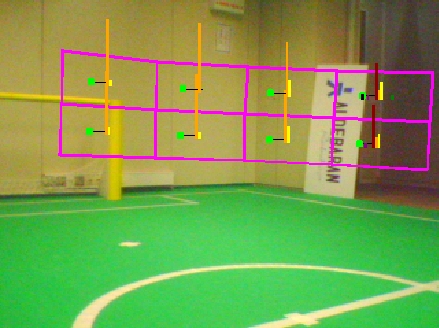}
            \label{fig:SymeterieTiles01b}
        }
        \subfigure[Real robot view of own goal.]
        {
            \includegraphics[width=0.3\textwidth]{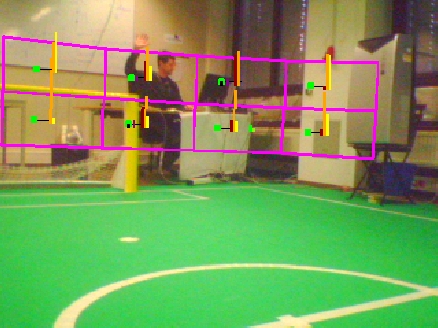}
            \label{fig:SymeterieTiles02b}
        }
        \subfigure[Simulated environment.]
        {
            \includegraphics[width=0.3\textwidth]{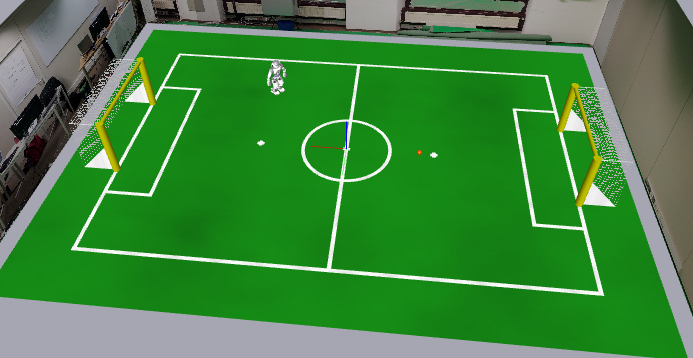}
            \label{fig:reflectionA_Sim}
        }
        \subfigure[Simulated robot view of opponent goal.]
        {
            \includegraphics[width=0.3\textwidth]{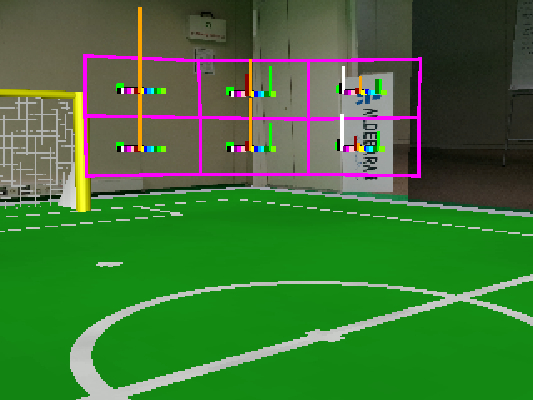}
            \label{fig:SymeterieTiles01b}
        }
        \subfigure[Simulated robot view of own goal.]
        {
            \includegraphics[width=0.3\textwidth]{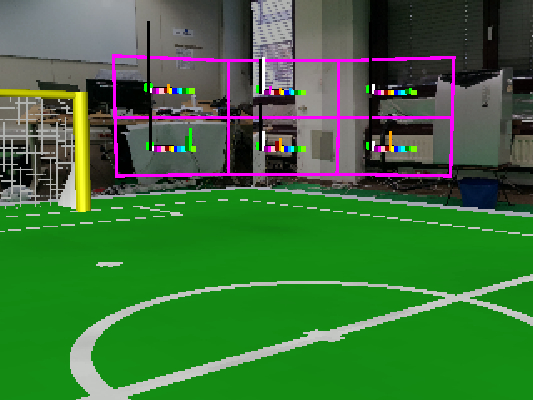}
            \label{fig:SymeterieTiles02b}
        }
        \caption{These figures present the scene perception with a real and simulated robot. The tiles projected depend on the robot's locations in (c).}
        \label{fig:SymetricEnvironment}
\end{figure}

\subsection{Non-Static Features on the Playing Field} \label{sec:Introduction:NonStaticFeatures}
The \textit{non-static features strategy} uses objects like the game ball or other detected objects in addition to features on the symmetric playing field. For example Fig.~\ref{fig:reflectionA_Sim} shows a single robot and a game ball on the playing field. Fig.~\ref{fig:TwoPossiblePoses} show the two possible poses. The robot can determine a correct pose by knowing the ball position.
This idea can be used with a single robot by tracking all objects of interest, but it works optimally with team communication, because  perceived ball and robot locations are shared.
This approach is therefore similar to multi-robot localization \cite{Fox2000}. 
The system starts to fail if the team has only one player left and this robot falls, because the stability of the non-static features cannot be guaranteed after a certain period of time. 
Some teams (e.g. the Dortmund Devils) have already integrated non-static features into their Kalman-Filter based localization~\cite{Czarnetzki2010}.

\subsection{Features Beyond the Playing Field} \label{sec:Introduction:NewFeatures}
Similar to humans, this strategy uses features beyond the symmetric playing field.
The idea is to identify outstanding features in the background and map them using  \textit{Self Localization and Mapping} (SLAM) approaches. 
Such techniques are widely used in robot localization. For instance, Anati~\cite{Anati2012} trained an object-detection algorithm to recognise, among others, clocks and trash cans in order to determine a robot's location within a train station.
State of the art object detection algorithm are often based on interest points, i.e., salient image regions and a descriptor. For example the most popular interest points are Lowes' SIFT~\cite{Loy2006} and Bays' SURF~\cite{Bay2006}. 
Because of the limited computational power in RoboCup SPL, it is impossible to use such object detection algorithm during a soccer game. 
As an alternative, Anderson~\cite{Anderson2012} presented a simplified 1D SURF descriptor to map backgrounds. 
In~\cite{Bader2012b} Bader et al. use colour histograms as descriptors. However, they did not propose a reliable strategy for matching those descriptors over time.

The challenges for any strategy which uses the background for localization can be summarized by:
\begin{itemize}
\item \textbf{Training}:
When should the background be trained?
\item \textbf{Matching}:
When should it be used for matching?
\end{itemize}
This work proposes the use of colour histogram based image features and a strategy for finding a balance between learning and training those features. 
It might be said that other features, e.g image gradient information, could also be used for mapping, but for the purposes of this study, only colour histograms were implemented.

\section{Room-Awareness} \label{sec:RoomAwareness}
The  technique proposed here improves existing particle filter based self-localization algorithms by allowing the robot's BC to change internal localization hypotheses. Fig.~\ref{fig:TheRoomAwarenessModule} shows the integration of the room-awareness module and the control channels to an the self-localization algorithm used (green). 

The external placement of the room-awareness module and the BC as the decision making unit allow for further integration of other techniques, e.g. non-static features like the game ball, without changing a well established localization technique. 
\\The room-awareness approach requires initial knowledge, such as the robot's pose or a previously learned visual background model.
The following sections describe the sub-modules needed in order to realise the approach.

\subsection{Perceived Colour Histograms and Background Tiles}

\begin{figure}[h]
  \centering
  \def\svgwidth{0.8\columnwidth}
 \executeiffilenewer{backgroundTilesMatlab.svg}{backgroundTilesMatlab.pdf}{inkscape -z -D --file=backgroundTilesMatlab.svg  --export-pdf=backgroundTilesMatlab.pdf --export-latex}%
 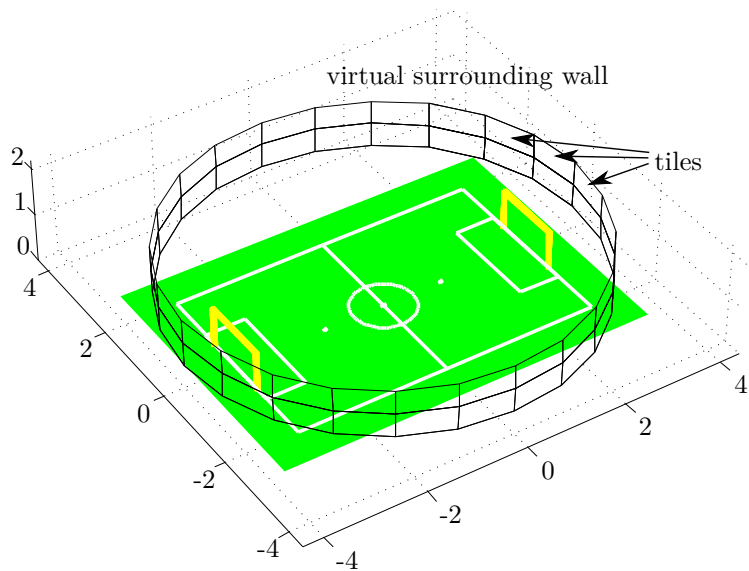%

  \label{fig:BackgroundTilesMatlab}
  \caption{Surrounding wall of tiles in multiple rows and columns in the shape of a cylinder.}
\end{figure}

\begin{figure}
  \centering
  \def\svgwidth{0.95\columnwidth}
 \executeiffilenewer{colours_histogram.svg}{colours_histogram.pdf}{inkscape -z -D --file=colours_histogram.svg  --export-pdf=colours_histogram.pdf --export-latex}%
 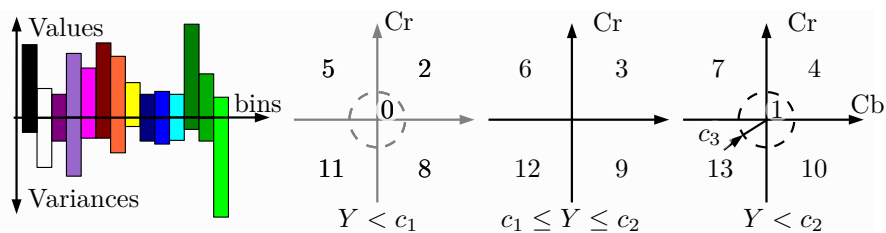%

  \label{fig:ColoursHistogram}
  \caption{A color histogram and bin values are drawn upwards, and the related variances downwards. Such histograms are used for modelling the background. The bins are defined by the axis sign and three thresholds $c_1$, $c_2$ and $c_3$, as drawn on the three YCrCb colour space slices.}
\end{figure}
The \textit{perceived colour histograms} are linked to tiles on a virtual surrounding, see Fig~\ref{fig:BackgroundTilesMatlab}. The cylinder-shaped virtual wall is composed of multiple rows and columns of rectangular tiles.
Figure \ref{fig:ColoursHistogram} shows an example of a colour histogram used to augment the robot's views in all other figures.
Colour bin values are drawn upwards and variance of each bin is drawn downwards.
Due to symmetry, the back-projected tiles appear equal for all reflected poses on the camera image, see Fig.~\ref{fig:SymetricEnvironment}(b,c,e,f).
The color histograms linked to the \textit{background tiles} can be used to disambiguate the viewing direction. 
Perceived histograms are drawn at the bottom left, outside the background model in Fig.~\ref{fig:BackgroundModel}, without variance.

\subsection{The Background Model}
The \textit{background model} is trained online with perceived histograms by using a moving average update strategy.
To stabilize the model, tiles which are blocked by other robots or observed from a certain viewing angle are ignored.
The moving average update strategy allows for the computation of variance for each colour histogram bin in order to detect unstable areas.
The equations for computing the moving average $\mu$ and variance $\sigma$ are shown in Eq.~(\ref{eq:MovingAverage})~and~(\ref{eq:MovingAverageCov}), but if a tile is seen for the first time, the perception is copied. 
Increasing $N$ leads to a more stable model but to a lower rate of adaptation to environmental changes.
\begin{equation}
    \label{eq:MovingAverage}
\mu_{new} = \dfrac{N \mu_{last} + x_{mess}}{N+1}
\end{equation} 
\begin{equation}
    \label{eq:MovingAverageCov}
\sigma^2_{new} = \dfrac{1}{N+1}\left(N \sigma^2_{last} + \dfrac{N}{N+1} * (\mu_{last}-x_{mess})^2 \right)
\end{equation} 
The training can be interrupted by the robot's BC to avoid learning from incorrect models, as when the robot falls or during a penalty in the soccer game. A trained background model with colour histograms around the playing field is shown in Fig.~\ref{fig:BackgroundModel}. The two circles of histograms indicate the two rows of tiles cylindrically arranged around the playing field.

\subsection{The Background Evaluation}
\begin{figure}
\centering 
        \raisebox{-0.5\height}{
        \subfigure[Background model and background particles drawn on the robot's internal world view around the playing field. Particles from the self-localization are drawn in gray on the playing field.]
        {
			
			\includegraphics[width=0.6\textwidth]{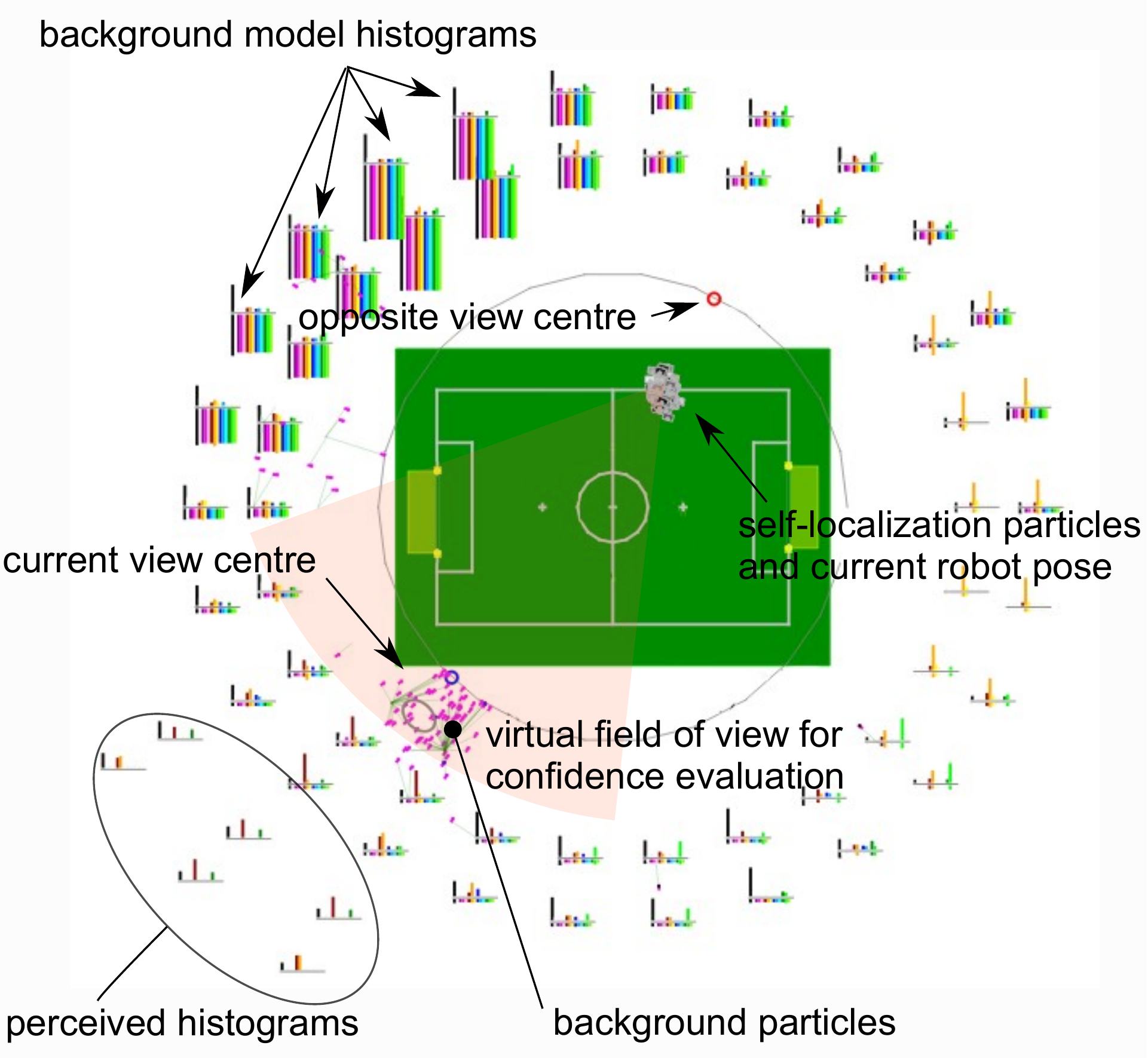}
            \label{fig:BackgroundModel}
        }}
        \raisebox{-0.5\height}{
        \subfigure[Camera image with particles and tiles with colour histograms. Detected landmarks like field lines and the goal posts are drawn as overlay.]
        {
            \includegraphics[width=0.30\textwidth]{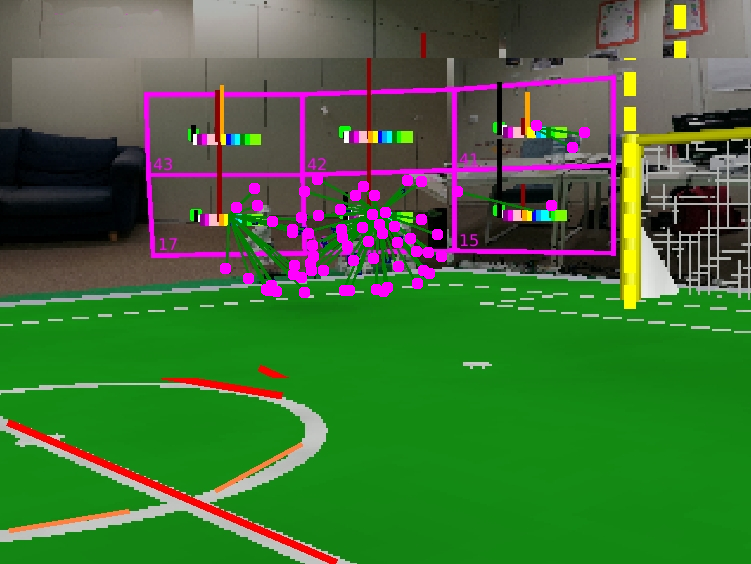}
            \label{fig:RoomAwarenessOnImage}
        }}
\caption{The robot's internal world view and the related camera image.}
\label{fig:RoomAwarenessDebug}
\end{figure}

This sub-module uses the robot pose information and the perceived colour histograms to estimate the current viewing direction based on the background model. 
The estimation is done by using a particle-filter where each particle describes a viewpoint hypothesis on the cylindrically modelled wall.
The particle weights are computed by comparing the colour histograms of the perceived tiles with model histograms of viewpoints estimated by a specific particle. 
All particles are updated by linearising the current rotational motion at the current view center plus some additional white noise to compensate for model discrepancy. 
New particles are injected at the current best matching position to prevent the filter from getting stuck in local minima.
Fig.~\ref{fig:BackgroundModel} indicates particles with magenta dots around the playing field and the detected cluster center is drawn as a gray ellipse. The same particles are visualized in the current perceived image, Fig.~\ref{fig:RoomAwarenessOnImage} (black).

\subsection{Confidence Values}
\begin{figure}
	\centering 
    \includegraphics[width=0.7\textwidth]{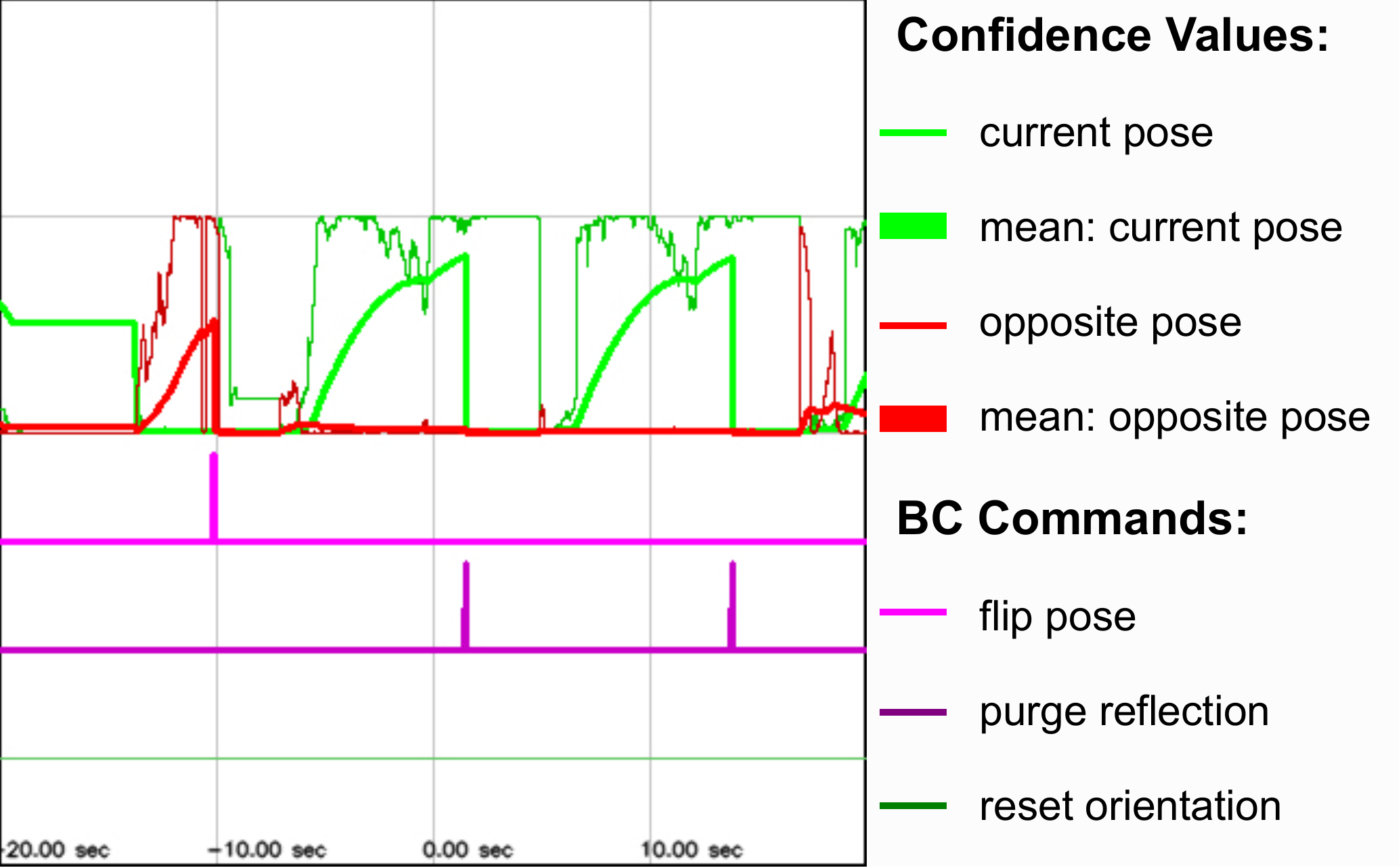}
\caption{History of past confidence values and BC's command signals. The BC recently triggered a \textit{flip} because the robot was placed incorrectly. This was followed by \textit{purge reflection} because the awareness module indicated a high current pose confidence over the reflected pose confidence.}
\label{fig:RoomAwarenessConfidenceValues}
\end{figure}

Multiple confidence values are computed by counting the number of background particles within certain angle ranges.
Confidence for the current pose is computed by counting the number of particles within a virtual field of view around the current view center, shown in Fig.~\ref{fig:BackgroundModel}.
The opposite pose confidence is computed similarly but with the opposite view center.
A moving average algorithm uses the previous values to suppress incorrect estimates and to assure a smoother confidence value. Fig.~\ref{fig:RoomAwarenessConfidenceValues} shows the previous measurements as signals over time, together with the BC command signals.

\section{Results} \label{sec:Results}
\begin{figure}[h]
\centering 
		\raisebox{-0.5\height}{
        \subfigure[All of the self-localization particles are on the incorrect pose.]
        {
			\includegraphics[width=0.4\textwidth]{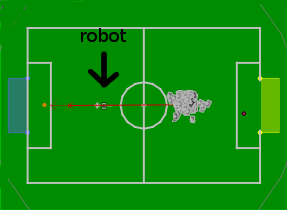}
            \label{fig:results_field_flip}
        }}
		\raisebox{-0.5\height}{        
        \subfigure[Most of the self-localization particles are on the correct pose.]
        {
            \includegraphics[width=0.4\textwidth]{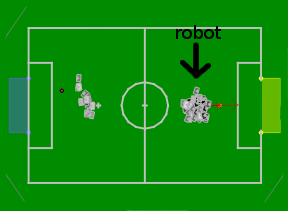}
            \label{fig:results_field_purge}
        }}
        
		\raisebox{-0.5\height}{
        \subfigure[A \textit{flip} solved the problem of an incorrect pose. Data was recorded on a real robot.]
        {
			\includegraphics[width=0.4\textwidth]{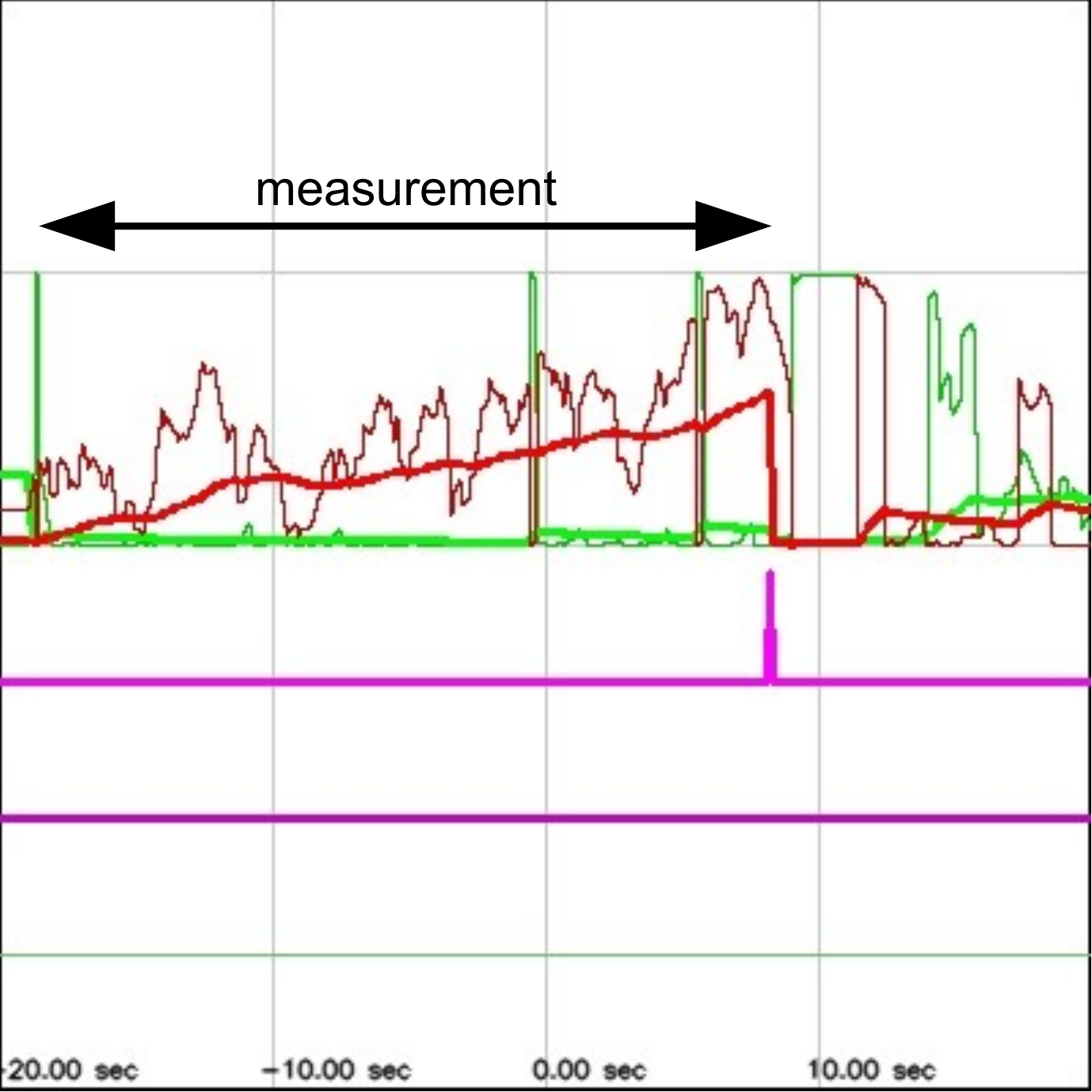}
            \label{fig:results_image_flip}
        }}        
		\raisebox{-0.5\height}{
        \subfigure[A \textit{purge} optimised the distribution. Data was recorded on a simulated robot.]
        {
            \includegraphics[width=0.40\textwidth]{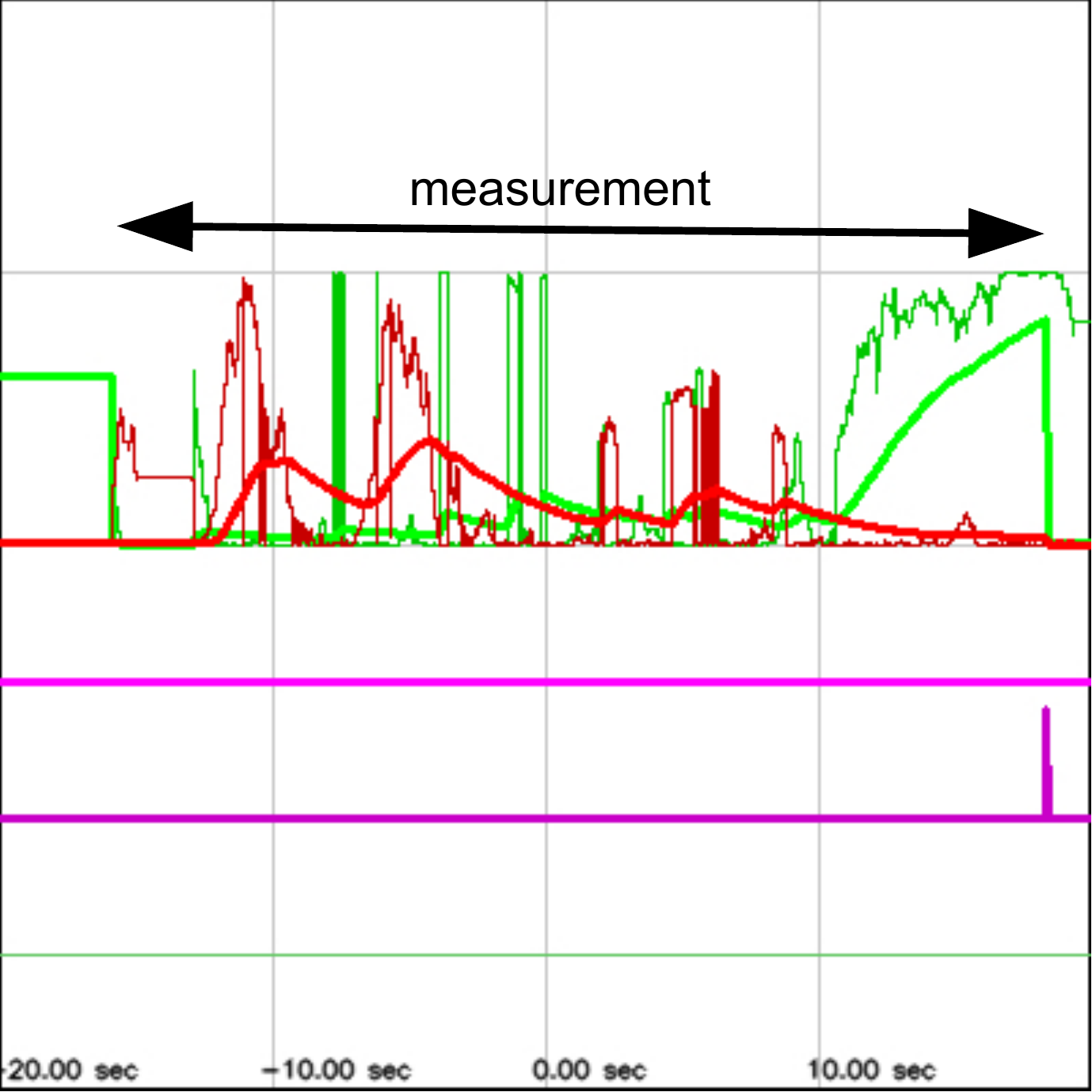}
            \label{fig:results_image_purge}
        }}
\caption{Two  cases were observed in which the system optimised the particle distribution after an incorrect initialization on a simulated robot.}
\label{fig:results_oszi}
\end{figure}

Two test scenarios were used on a simulated and real robot to measure this approach's improvement on a system without a room-awareness module. For both tests the robot was placed  on the default soccer start position next to the playing field, but the particle-filter based self-localization was initialized with the opposite "incorrect" pose, and sides were switched  after half the trials. A system which uses only the symmetric playing field for localization would converge only by accident to the correct pose and would normally fail. In the \textit{first test} the robot was allowed  to move only its head. In the \textit{second test} the robot had to walk from one penalty position to the other and vice versa, based on its own localisation. During all of the trials the time it took for the BC to trigger a control command to optimize or correct the self-localization's particle distribution was measured. 
Table~\ref{tab:ResultsReal} shows the number of trials and the average time it took for the BC to create a signal. 
\begin{table}[h]
\centering
\small
\begin{tabular}{|l|l|l|l|l|l| }
\hline
\textbf{Simulated Robot}        & trials & flip & purge & failed \\ \hline\hline
not moving  & 20 trials & 70\% & 15\% & 15\% \\ 
average time to a signal  & 18.1 sec & 12.9 sec & 57.5 sec & $>$ 200 sec \\ \hline
moving & 10 trials & 80\% & 20\% & 0\% \\ 
average time to a signal  & 20.2 sec & 17.1 sec & 34.5 sec & - \\ \hline \hline
\textbf{Real Robot}        & trials & flip & purge & failed \\ \hline\hline
not moving  & 20 trials & 90\% & 0\% & 10\% \\ 
average time to a signal  & 23.1 sec & 23.1 sec & - & 
$>$ 200 sec \\ \hline
moving & 10 trials & 100\% & 0\% & 0\% \\ 
average time to a signal  & 33.5 sec & 33.5 & - &  - \\ \hline
\end{tabular}
\caption{Results from a simulated and real robot}
\label{tab:ResultsReal}
\end{table}
One can see that the system tends to fail up to 20\% if the robot is not moving.
This happens because the background is trained online and the wrong background is assumed as correct over a certain time.
Two ways in which the BC corrects the self-localization distribution were also observable, Fig.~\ref{fig:results_oszi}. 
If the particles are on the wrong pose, the system triggers a flip. A purge was called for if the filter accidentally formed a correct growing particle cluster.
The reason why the system on the real robot never triggered a purge in these tests is because, for both the simulation and the real robot, the room-awareness parameters were fixed to generate comparable results. 
On the real robot smoother confidence values were experienced than on the simulated, visible in Fig.~\ref{fig:results_image_flip}, for which the noisier and smoother image was held responsible. 
Overall, it was shown that the proposed room-awareness module improves an existing self-localization algorithm. 
\section{Conclusion} \label{sec:Conclusion}
In this paper, a room-awareness module inspired by the results of psychological studies was presented which mimics a human belief of current pose and triggers a so-called spontaneous reorientation.
Results of tests with a humanoid robot on a RoboCup SPL playing field supported the initial belief in recognition of an incorrect estimated pose due to symmetric environments by mapping the surrounding environment with colour histograms. 
In addition, a less-cluttered particle distribution in the self-localization used was generated during normal operation by allowing the BC to interfere with the self-localization by selectively moving or removing particle clusters caused by a symmetric environment. 
However, multiple issues remain open, e.g. an optimal count of color histograms, a different feature for modelling environments such as gradient information, the  virtual wall shape used, and the implementation of an optimal search pattern to force the robot to look at areas with the most distinctive background.
\section*{Acknowledgments}
The research leading to these results has received funding from the Austrian Science Foundation under grant agreement No.~835735~(TransitBuddy)
\bibliography{refs}
\end{document}

%% file: backgroundTilesMatlab.pdf_tex
\begingroup%
  \makeatletter%
  \providecommand\color[2][]{%
    \errmessage{(Inkscape) Color is used for the text in Inkscape, but the package 'color.sty' is not loaded}%
    \renewcommand\color[2][]{}%
  }%
  \providecommand\transparent[1]{%
    \errmessage{(Inkscape) Transparency is used (non-zero) for the text in Inkscape, but the package 'transparent.sty' is not loaded}%
    \renewcommand\transparent[1]{}%
  }%
  \providecommand\rotatebox[2]{#2}%
  \ifx\svgwidth\undefined%
    \setlength{\unitlength}{410.73213073bp}%
    \ifx\svgscale\undefined%
      \relax%
    \else%
      \setlength{\unitlength}{\unitlength * \real{\svgscale}}%
    \fi%
  \else%
    \setlength{\unitlength}{\svgwidth}%
  \fi%
  \global\let\svgwidth\undefined%
  \global\let\svgscale\undefined%
  \makeatother%
  \begin{picture}(1,0.77603594)%
    \put(0,0){\includegraphics[width=\unitlength]{backgroundTilesMatlab.pdf}}%
    \put(0.44050632,0){\makebox(0,0)[lb]{\smash{-4}}}%
    \put(0.58074371,0.06393218){\makebox(0,0)[lb]{\smash{-2}}}%
    \put(0.71891892,0.12373758){\makebox(0,0)[lb]{\smash{0}}}%
    \put(0.85090519,0.18354298){\makebox(0,0)[lb]{\smash{2}}}%
    \put(0.98082928,0.24128865){\makebox(0,0)[lb]{\smash{4}}}%
    \put(0.33326611,0.01649737){\makebox(0,0)[lb]{\smash{-4}}}%
    \put(0.24665005,0.11136455){\makebox(0,0)[lb]{\smash{-2}}}%
    \put(0.17034484,0.20210496){\makebox(0,0)[lb]{\smash{0}}}%
    \put(0.09197746,0.29078319){\makebox(0,0)[lb]{\smash{2}}}%
    \put(0.01567226,0.37327735){\makebox(0,0)[lb]{\smash{4}}}%
    \put(0.00742114,0.43102058){\makebox(0,0)[lb]{\smash{0}}}%
    \put(0.00329704,0.49082841){\makebox(0,0)[lb]{\smash{1}}}%
    \put(-0.00082779,0.55269842){\makebox(0,0)[lb]{\smash{2}}}%
    \put(0.42712886,0.66288904){\color[rgb]{0,0,0}\makebox(0,0)[lb]{\smash{virtual surrounding wall
}}}%
    \put(0.87233424,0.54590076){\color[rgb]{0,0,0}\makebox(0,0)[lb]{\smash{tiles}}}%
  \end{picture}%
\endgroup%

%% file: colours_histogram.pdf_tex
\begingroup%
  \makeatletter%
  \providecommand\color[2][]{%
    \errmessage{(Inkscape) Color is used for the text in Inkscape, but the package 'color.sty' is not loaded}%
    \renewcommand\color[2][]{}%
  }%
  \providecommand\transparent[1]{%
    \errmessage{(Inkscape) Transparency is used (non-zero) for the text in Inkscape, but the package 'transparent.sty' is not loaded}%
    \renewcommand\transparent[1]{}%
  }%
  \providecommand\rotatebox[2]{#2}%
  \ifx\svgwidth\undefined%
    \setlength{\unitlength}{1133.8bp}%
    \ifx\svgscale\undefined%
      \relax%
    \else%
      \setlength{\unitlength}{\unitlength * \real{\svgscale}}%
    \fi%
  \else%
    \setlength{\unitlength}{\svgwidth}%
  \fi%
  \global\let\svgwidth\undefined%
  \global\let\svgscale\undefined%
  \makeatother%
  \begin{picture}(1,0.25401305)%
    \put(0,0){\includegraphics[width=\unitlength]{colours_histogram.pdf}}%
    \put(0.44046569,0.23169871){\makebox(0,0)[lb]{\smash{Cr}}}%
    \put(0.67904392,0.23011113){\makebox(0,0)[lb]{\smash{Cr}}}%
    \put(0.269267,0.13633536){\makebox(0,0)[lb]{\smash{bins}}}%
    \put(0.90139354,0.23055213){\makebox(0,0)[lb]{\smash{Cr}}}%
    \put(0.4338491,0.00402587){\makebox(0,0)[b]{\smash{$Y<c_1$}}}%
    \put(0.85071953,0.00394095){\makebox(0,0)[lb]{\smash{$Y<c_2$}}}%
    \put(0.36937732,0.17701535){\makebox(0,0)[lb]{\smash{5}}}%
    \put(0.47936144,0.17701535){\makebox(0,0)[lb]{\smash{2}}}%
    \put(0.36505557,0.06209208){\makebox(0,0)[lb]{\smash{11}}}%
    \put(0.47936144,0.06209208){\makebox(0,0)[lb]{\smash{8}}}%
    \put(0.43693773,0.1295643){\makebox(0,0)[lb]{\smash{0}}}%
    \put(0.59437291,0.17701535){\makebox(0,0)[lb]{\smash{6}}}%
    \put(0.70444523,0.17701535){\makebox(0,0)[lb]{\smash{3}}}%
    \put(0.58899277,0.06209208){\makebox(0,0)[lb]{\smash{12}}}%
    \put(0.70444523,0.06209208){\makebox(0,0)[lb]{\smash{9}}}%
    \put(0.81442935,0.17701535){\makebox(0,0)[lb]{\smash{7}}}%
    \put(0.92441348,0.17701535){\makebox(0,0)[lb]{\smash{4}}}%
    \put(0.80904922,0.06209208){\makebox(0,0)[lb]{\smash{13}}}%
    \put(0.91656377,0.06209208){\makebox(0,0)[lb]{\smash{10}}}%
    \put(0.88190157,0.1295643){\makebox(0,0)[lb]{\smash{1}}}%
    \put(0.36937732,0.17701535){\makebox(0,0)[lb]{\smash{5}}}%
    \put(0.47936144,0.17701535){\makebox(0,0)[lb]{\smash{2}}}%
    \put(0.36505557,0.06209208){\makebox(0,0)[lb]{\smash{11}}}%
    \put(0.47936144,0.06209208){\makebox(0,0)[lb]{\smash{8}}}%
    \put(0.43693773,0.1295643){\makebox(0,0)[lb]{\smash{0}}}%
    \put(0.03386841,0.2242018){\makebox(0,0)[lb]{\smash{Values}}}%
    \put(0.03369201,0.02716528){\makebox(0,0)[lb]{\smash{Variances}}}%
    \put(0.79918404,0.10330454){\makebox(0,0)[lb]{\smash{$c_3$}}}%
    \put(0.97337417,0.13603025){\makebox(0,0)[lb]{\smash{Cb}}}%
    \put(0.65383156,0.00422373){\color[rgb]{0,0,0}\makebox(0,0)[b]{\smash{$c_1 \leq Y \leq c_2$}}}%
  \end{picture}%
\endgroup%